# DEVELOPING ICC PROFILE USING GRAY LEVEL CONTROL IN OFFSET PRINTING PROCESS

Jaswinder Singh Dilawari[†] andDr.RavinderKhanna[††]
*dilawari.jaswinder@gmail.com*,ravikh_2006@yahoo.co.in
[†]Ph.D Research Scholar
Pacific Academy of Higher Education and Research University,Udaipur,Rajasthan,India
[††]Principal, Sachdeva Engineering College for Girls, Mohali, Punjab,India

**Summary:** In prepress department RGB image has to be converted to CMYK image. To control that amount of black, cyan, magenta and yellow has to be controlled by using color separation method. Graycolor separation method is selected to control the amounts of these colors because it increase the quality of printing also. A single printer used for printing the same image on different paper also results in different printed images. To remove this problem a different ICC profile based on gray level control is developedand a sheet offset printer is calibrated using that profile and a subjective evaluation shows satisfactory results for different quality papers.

## 1. Introduction

The initial image handled by prepress department is RGB image. That has to be converted into CMYK for further processing.This color conversion is today done by ICC-profiles. The profiles contain information about separation, black start, black width, total ink coverage.GCR (Gray Component Replacement) and UCR (Under Color Removal) are the two main color separation techniquesused to control the amounts of black, cyan, magenta and yellow needed to produce the different tones. Since black ink can replace equal amounts of cyan, magenta and yellow. To produce a similar tone, UCR and GCR replace equal amounts of cyan, magenta and yellow in neutral tones. GCR also replaces some CMY colors in tertiary colors [1][2][3]. GCR is extended term of UCR. The main difference between UCR and GCR is that UCR works only in the neutral dark areas and GCR works for entire tonal range. So in RGB to CMYK conversion UCR has been discarded. With the possibility of developing new ICC profiles the problem of creating separate ICC profiles arises. This can be overcome by limiting the number of ICC profiles and it will be made possible by calibrating the printing press. If gray balance is controlled in printing press then printing press is automatically calibrated. The appropriate gray value decides the tone of cyan, magenta and yellow colors[4]. It is affected by color sequence, ink trapping, press characteristics and dot gain. Proper gray balance ensures that a tone of appropriate cyan, magenta and yellow tint values is visually perceived as neutral gray. The human eye is most sensitive to change in neutral gray. As colour becomes more saturated and nearer to gray a small change will be detected by naked eyes. To balance level of gray colour sheet offset printers have been designed and calibrated so that the colour reproduction on different papers will be similar to permit the use of fewer ICC profiles[5][6][7][8]. In cold offset printers gray level is balanced for the adjustment of ink quality in printer. Whereas in sheet fed printers this ink quality is controlled by deciding upon target tone value increase or density levels. Even then same printing quality can't be achieved in every run. So there is need of calibration of sheet fed offset printers. Since there is no standardized way to calibrate a sheet-fed offset press, the possibility of producing an equal print quality in different print runs and printing presses is limited.Gray-balance control is one way to calibrate a printing press. The press will then always be set to print one standardized combination of CMY halftones in the same way. Hopefully, this will create a similarity between print runs, presses and to some extent paper grades, which provide the foundation for robust ICC-profiles.

## 2. Press calibration using Gray level control in robust ICC colour Management

To calibrate the press first the gray balance control in sheet offset is explored and the stability of ICC profile is assessed in gray balance controlled print runs. To set the inking level of printer, printer is set to print neutral CMY combinations on five wood free coated papers (ISO12467-2). If the press is set to this inking level every time so that result looks same then it is possible to use same ICC profile in every run. Separate ICC profile are created for each of five papers. To compare the different combinations of ICC profiles and papers, CIE LAB images and test charts were converted to CMYK, printed and then measured or evaluated visually. The deviation in ΔEunits was used as a measure of the stability[9][12][14].

The gray-balance control method can be utilized in sheet-fed offset. The printable colorgamuts and the neutrality in the three-color gray were sufficiently similar between papers and print runs to enable good gray-balance control to be achieved[10]. In sheet-fed offset printer gray-balance control should be used in combination with target density levels. The ICC-profiles created from the gray-balanced print run were found to be quite robust [11][15].By gray balance control method the printed results were similar irrespective of use of paper whether classic matte, dull silk or gloss paper.If an average ΔE* ≤ 3 and the worst ΔE ≤ 5 are considered to be good enough, one ICC-profile for coated (fine) paper ought to be sufficient, preferably created for a glossy paper.

## 3. Gray Balance Control in Sheet Fed Offset Printer

The gray balance control is evaluated by density measurement, spectral measuring and comparison of print profiles [16]. The paper used in testing is a glossy paper 130g/m$^2$.in printing a colour image and a neutral image is used as shown in figure1.

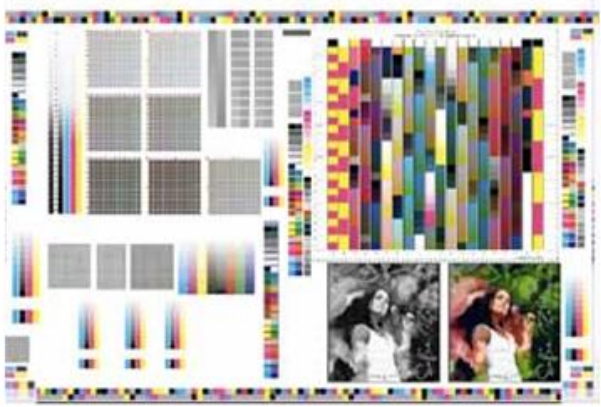

Figure 1: Neutral and colour image used for testing

The print tests, called Printing 1 (i.e. controlling by density) and Printing 2 (i.e. controlling by gray balance) were performed in two ways on the same day and Printing 3 (i.e. printing of the images separated by profiles based on Printing 1 and 2) was printed two months later.

### 3.1 Print, which is measure by density control

100 prints from the accepted prints were selected and measured by Techkon spectrophotometer that measures ΔE* [17]. Tern of these which have best match with ISO standard 12647-2:2004/Amd.1:2007(E), are selected and also again measured by Eyeone (X- Rite).

### 3.2 Print, which is controlled only by gray balance

The sheets which were needed by the printer to obtain an acceptable gray balance were collected and measured in order to check human eyes. 100 sheets from the accepted prints were randomly selected and measured (Techkon spectrophotometer, measuring of the ΔE). Ten of these 100 sheets, which had the best agreement with the ISO-standard 12647-2:2004/ Amd.1:2007(E), were also measured by EyeOne (X-Rite). The subjective evaluation of the printed images was done by paired comparison. Print ICC-profiles based on the best sheets were created. In subjective evaluation ten people are used [17][18][19]. The printer needed 26 sheets to finish the check. Ten sheets(A, B, C, D, E, F, G, H, I, J) in printing 2 have been evaluated. The subjective evaluation selected three best sheets H, I, J. In this, problem of density variation across the sheets is found. After sheet 26, the following sheets had ΔE ≈ 5.surpassing these facts sheet 27 has received the lowest numbers for the neutral image in the subjective evaluation and next lowest number of points for the colour image. The sheet 26 received highest number of points. It shows profiles based on the sheets are almost identical.

### 3.3 Print which contains the images separated by different profiles

The colour image and neutral image are separated using new ICC profiles developed based on profiles developed in printing 1 and 2. Results of this new ICC profile are evaluated subjectively and also compared with the same image separated by standardised ICC profile, ISO oated_v2_300_eci.icc.

## 4. Creation of profile Based on Printing by Density and Printing by Gray Balance

Best sheet from printing 1 are A and I and from printing2 are H, I, J. New profiles from these best sheets have been created and compared with each other and with ISO coated_v2_300_eci.icc. See figure 2. The profile based on Printing1 (controlling by gray balance) is more similarin size to ISOcoated_v2_300_eci.icc than the profile based on Printing 3. A new test form and a new control strip based on checking density and gray balance were created for Printing 5, see Figure 3. The test form contains the neutral and the color images separated with the *ISO coated_v2_300_eci.icc* (www.eci.org)[20][21][22][23] and the profiles based on Printing 1 and Printing 2. The new control strip is based on the Heidelberg strip used in the printing company. The printer had the possibility to see and visually check the gray balance across the sheet but he also had the possibility to measure the density.

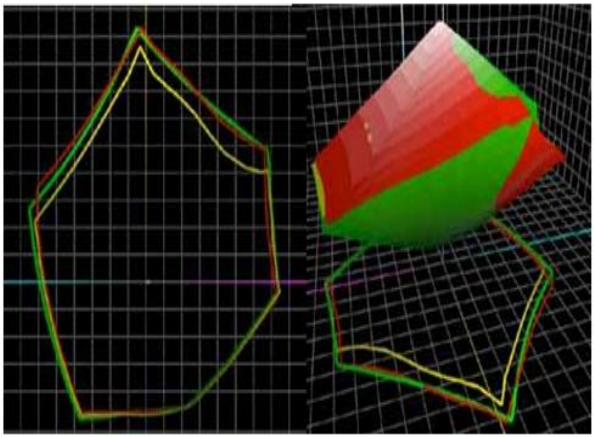

Figure2: The comparison between the profiles: the red profile is Printing 4, the yellow profile is Printing 3 and the green profile is ISOcoated_v2_300_eci.icc.profile.

A subjective comparison and evaluation was done between the printed images based on new IT8 test chart [24]. In the first evaluation step the printed images separated with a profile based on printing 3 and 4 were tested. In second evaluation the test people found out which of these two images is more similar to image that was separated using ISOcoated_v2_300_eci.icc.

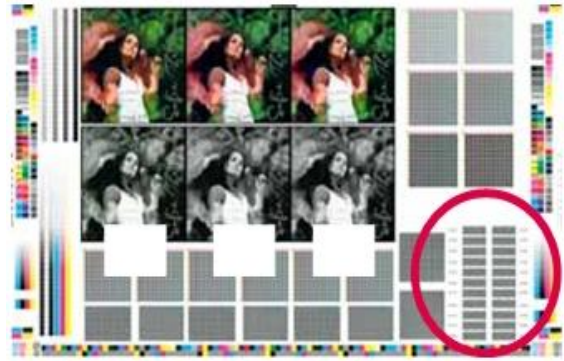

Figure3: The new test form for Printing 5, the red circle shows the hidden words inside the patches.

Evaluation resultsshowed that image separated using gray balance profile is more preferred than density control profile. However, all test people said that both images (neutral and color) from Printing 4 were most similar to the image, which were separated by ISOcoated_v2_300_eci.icc [25][26].

| Neutral image | Color image | Similarity with ISO |
|---|---|---|
| Printing 4 = 10 points | Printing 4 = 10 points | Printing 4 = 10 points |
| Printing 3 = 0 point | Printing 3 = 0 points | Printing 3 = 0 points |

Table1: Subjective evaluation result and printer4 got highest points

## Result

Tests made using the currently available software (Adobe Photoshop - image processing, ProfileMaker - profile creation, Print-Open - profile creation)for separation of color images GCR color separation is more preferred rather than UCR because GCR more concerns with gray component replacement and even printers are having same ICC profile even then same image printed is different on different paper. To remove this problem gray balance print run ICC profiles has been created. Printing using this profile is same on dull, gloss and dull silk paper whereas for coated paper it is just satisfactory. In actual printer can be calibrated to produce same image irrespective of type of paper and paper quality. But gray balance control strip is proved to be good instrument and it is tested on sheet fed offset printer. This gray balance control also increased the quality of printer.

## References

[1] Sharma Abhay ,"Understanding Color Management" Thomson Delmar Learning, USA ,2004

[2] SIS 2009, www.sis.se,Swedish Standards Institute

[3] Tidningsutgivarna2002 "Gråbalansstyrdtryckning" (Electronic PDF) <http://www.tu.se> 2003-03-05.

[4] Tidningsutgivarna 1990 "Färgpressen", Stockholm, Sweden

[5]Toledo-bend 2007; Ishihara,www.toledo-bend.com/colorblind/Ishihara.asp

[6] UGRA2009, www.ugra.ch, Swiss Centre of Competence for Media and Printing Technology

[7] Wallner Dawn 2000 " Building ICC-Profiles - the Mechanics and Engineering" dawn.wallner@yahoo.com

[8] Yule J. A. C. 1940 "Four- Color processes and the Black Printer" Journal of the Optical Society of America, Vol. 30, August 1940

[9] Yule J. A. C. 1967 "Principles of Color Reproduction" Wiley, New York

[10] Åman A, Lind J. 2004. "Bildtypsanpassadprofil till en skrivare - utvärdering and analysHögskolanDalarna

[11] Standardization in Offset Printing, "Heidelberger Druckmaschinen AG",www.heidelberger.com

[12] S. Kamata, M. Nimmi, and E. Kawaguchi, "A gray image compression using Hilbert scan", Proc. International Conf. on Pattern Recognition(ICPR)(IEEE Computer Press, Los Alamitos, CA, 1996, Vol. 3, pp.905–909.

[13] Sharma, G. and Rodr´ıguez-Pardo, C. E., "The dark side of CIELAB," in [Proc. SPIE: Color Imaging XVII:
Displaying, Hardcopy, Processing, and Applications], **8292**, 8292–12,1–9 (Jan. 2012).

[14] Gaurav Sharma, Robert P. Loce, Steven J. Harrington, Y. (Juliet) Zhang,"Illuminant Multiplexed Imaging: Special Effects using GCR" IS&T/SID Eleventh Color Imaging Conference

[15] M. J. Vrhel and H. J. Trussell, "Color Device Calibration: A Mathematical Formulation," *IEEE Transactions on Image Processing*, Vol. 8, No. 12, Dec. 1999.

[16] M. C. Stone, W. B. Cowan, and J. C. Beatty, "Color Gamut Mapping and the Printing of Digital Color Images," ACM Transactions on Graphics, Vol. 7, No. 4, Oct. 1988.

[17] M. Xia, E. Saber, G. Sharma, and A. M. Tekalp, "End-to-End Color Printer Calibration by Total Least Squares Regression," IEEE Trans. on Image Proc., vol. 8, no. 5, pp. 700-716, May 1999.

[18] M. J. Vrhel, H. J. Trussell, " Color Printer Characterization in MATLAB" IEEE ICIP 2002(p457-460)

[19] Gaurav Sharma," Digital Color Imaging Handbook" Xerox Corporation, Webster, New York, USA

[20] Raja Bala and GauravSharma,"System Optimization in Digital Color Imaging"

[21] R. Bala, "Device characterization," in Digital Color Imaging Handbook, G.Sharma, Ed. Boca Raton, FL: CRC, 2003, ch. 5.

[22] R. Balasubramanian and R. Eschbach, "Reducing multi-separation color moiré via a variable undercolor removal and gray-component replacement strategy," J. Imaging Sci. Technol., vol. 45, no. 2, pp. 152–160, Mar./Apr. 2001.

[23] G. Sharma, "Target-less scanner color calibration,'' J. Imaging Sci. and Tech.,vol. 44, no. 4, pp. 301–307, Jul./Aug. 2000.

[24] I. C. Consortium, "Icc.1:2004-10," in http ://www.color.org/icc specs2.html, 2004.

[25] H.R. Kang, *Color Technology for Electronic Devices.* Bellingham WA: SPIE Press, 1997.

[26] K. T. Knox, Integrating cavity effect in scanners, in Proc. IS&T/OSA Optics and Imaging in the Information Age, IS&T, Springfield, VA, 1996, pp. 83–86

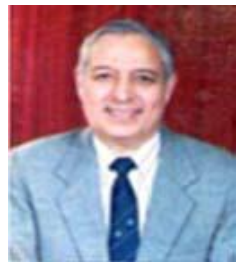
Born in 1948, Dr.RavinderKhannaGraduated in Electrical Engineeringfrom Indian Institute of Technology(IIT) Dehli in 1970 and Completed hisMasters and Ph.D degree in Eletronicsand Communications Engineering fromthe same Institute in 1981 and 1990respectively. He worked as an ElectronicsEngineer in Indian Defense Forces for 24Years where he was involved in teaching,research and project magement of someof the high tech weapon systems. Since 1996 he has full timeSwitched to academics. he has worked in many premiere technicalinstitute in india and abroad. Currently he is the Principal ofSachdeva Engineering College for Girls, Mohali, Punjab (India).He is active in the general area of Computer Networks, ImageProcessing and Natural Language Processing.

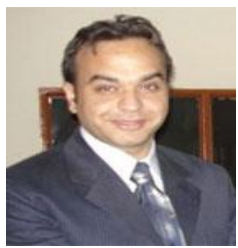
**Jaswinder Singh Dilawari** is pursuing Ph.D in the Dept of Computer Sci at Pacific Academy of Higher Education and Research University,Udaipur,Rajasthan ,INDIA and working as an Associate Professor ,Geeta Engineering College,Panipat,Haryana ,India .He has teaching experience of 12 years .His area of interest includes Computer Graphics, Computer Architecture ,Software Engineering ,Fuzzy Logic  and Artificial Intelligence .He is life member of  Indian Society for Technical Education (ISTE)